\pdfoutput=1

\documentclass[11pt]{article}

\usepackage[final]{acl}

\usepackage{times}
\usepackage{latexsym}

\usepackage[T1]{fontenc}

\usepackage[utf8]{inputenc}

\usepackage{microtype}

\usepackage{inconsolata}

\usepackage{graphicx}

\usepackage{algorithm}
\usepackage{algorithmicx}
\usepackage{algpseudocode}

\usepackage{amsmath}  

\usepackage{enumitem} 
\usepackage{multirow} 
\usepackage{multicol} 

\usepackage[capitalise]{cleveref}

\usepackage[colorinlistoftodos,prependcaption]{todonotes}

\title{Neural Search Space in Gboard Decoder}

\author{Yanxiang Zhang\thanks{Equal contribution.}, Yuanbo Zhang$^*$, Haicheng Sun, Yun Wang, Billy Dou, \\
        \textbf{Gary Sivek, Shumin Zhai} \\
        Google Inc \\
        \texttt{{zhangyx,zyb,haicsun,wyun,billydou,gsivek,zhai}@google.com}}

\begin{document}
\maketitle

\begin{abstract}

Gboard Decoder produces suggestions by looking for paths that best match input touch points on the context aware search space, which is backed by the language Finite State Transducers (FST). The language FST is currently an N-gram language model (LM).  However, N-gram LMs, limited in context length, are known to have sparsity problem under device model size constraint. In this paper, we propose \textbf{Neural Search Space} which substitutes the N-gram LM with a Neural Network LM (NN-LM) and dynamically constructs the search space during decoding. Specifically, we integrate the long range context awareness of NN-LM into the search space by converting its outputs given context, into the language FST at runtime. This involves language FST structure redesign, pruning strategy tuning, and data structure optimizations. Online experiments demonstrate improved quality results, reducing Words Modified Ratio by [0.26\%, 1.19\%] on various locales with acceptable latency increases. This work opens new avenues for further improving keyboard decoding quality by enhancing neural LM more directly.

\end{abstract}
\section{Introduction}

Gboard is a statistical-decoding-based keyboard on mobile devices developed by Google. Statistical decoding is far more necessary than one might think due to the error-prone process of “fat finger” touch input on small screens. According to \citet{azenkot2012touch}, the per-letter error rate is around 8\%-9\% without decoding. With decoding, typos such as substitutions (due to the proximity of two keys or cognitive misspellings), omissions, insertions, and transpositions could be automatically corrected by the key-correction and (word) auto-correction functions in the Gboard decoder, leading to an error-tolerant user experience. Powered by language models (LM), the Gboard decoder also provides rich functionalities such as word completion, post correction, next word prediction, smart compose (in-line predictions) to further save users' physical input effort.

\begin{figure*}
    \centering
    \includegraphics[width=15cm]{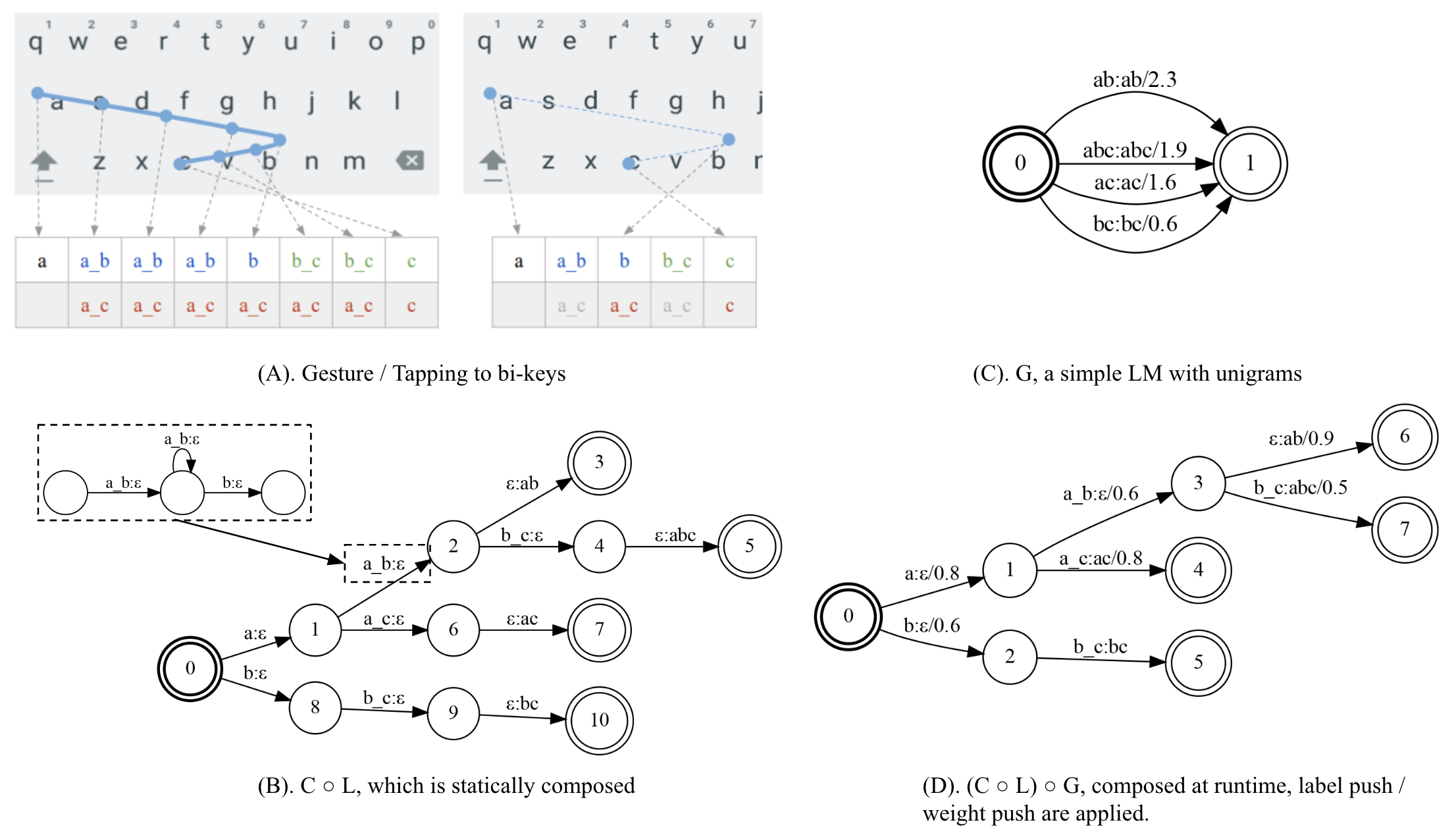}
    \caption{Build search space by composing $(C \circ L) \circ G$}
    \label{fig:compose}
    \vspace{-0.5cm}
\end{figure*}

The decoding process involves two phases: building search space (decoder graph), and performing beam search within the space based on user touch inputs. Gboard decoder utilizes context, a lexicon and language transducers - the familiar $C \circ L \circ G$ composition \cite{ouyang2017mobile, hellsten2017transliterated} - to construct the search space. $C$ is a bi-key key to key transducer while $L$ is a key to word transducer, $C$ and $L$ are statically composed together offline since the size is small. \cref{fig:compose}-A illustrates how gesture typing and tap typing inputs are converted into bi-keys and \cref{fig:compose}-B illustrates a composed $C \circ L$ targeting four words. Before this work, $G$ is a N-gram language FST containing 64k words for n-grams and 170k words for uni-grams. Composition between $(C \circ L)$ and $G$ are dynamically conducted due to the large size of $G$. \cref{fig:compose}-C shows a simple $G$ containing only four words, and \cref{fig:compose}-D illustrates a composed $(C \circ L) \circ G$, which is similar to $(C \circ L)$ but with weights achieved by using the look-ahead composition filters proposed by \citet{allauzen2009generalized, allauzen2011filter}.

In practice, the whole search space of $(C \circ L) \circ G$ like \cref{fig:compose}-D can't be fully expanded due to its huge size. Only states which are close to users’ bi-key inputs will be expanded. Specifically, the states are pruned based on the combination of LM scores and spatial scores in the decoder graph while the user is typing.

In this work, the N-gram LM is replaced by the NN-LM. Due to the framework complexity brought by the rich functionalities, we propose a runtime conversion solution to minimize the framework change. We call the search space built on the NN-LM the \textbf{Neural Search Space (NSS)}, and the original one the N-gram Search Space.

\cref{alg:init_search_space} and \cref{alg:extend_search_space} describe the changes of NSS in initializing and extending the search space at a high level, in which the codes in red are for Neural Search Space and the codes in blue are for N-gram Search Space.


\begin{algorithm}
    \caption{Initialize Search Space}
    \label{alg:init_search_space}
    \begin{algorithmic}
        \Ensure{$C_{input}$}\Comment{What users have committed}
        \Require{$State_{SS}$}\Comment{Initial states in search space}
        \State $S_{C \circ L}\gets0$
        \If{\textcolor{blue}{N-gram LM}}
        \State \textcolor{blue}{$G\gets{G_{ngram}}$}
        \State \textcolor{blue}{$S_G\gets{FindState(C_{input}, G_{ngram})}$}
        \ElsIf{\textcolor{red}{NN-LM}}
        \State \textcolor{red}{$G_{NLM}\gets{UpdateLM(NLM, C_{input})}$}
        \State \textcolor{red}{$G\gets{G_{NLM}}$}
        \State \textcolor{red}{$S_G\gets0$}
        \EndIf{}
        \State $State_{SS}\gets{Compose(C \circ L, G, S_{C \circ L}, S_G)}$
    \end{algorithmic}
\end{algorithm}

\cref{alg:init_search_space} is called before users start to type a new word, for example, when users open the keyboard for an editor box or when users commit a word by tapping on space. \cref{alg:extend_search_space} is called when users are typing a word; for tap-typing, it will be called upon each key tap.

NSS has minor changes in both algorithms, In \cref{alg:init_search_space}, rather than finding a specific state in a static N-gram LM given context, $UpdateLM$ inserts the next words and corresponding probabilities given context at the start state of $G_{NLM}$ as arcs, thus the start state is always 0. In \cref{alg:extend_search_space}, $ExtendLM$ additionally extends the $G_{NLM}$ which aims to handle the multi-word problem discussed later.

\begin{algorithm}
    \caption{Extend Search Space During typing}
    \label{alg:extend_search_space}
    \begin{algorithmic}
        \Ensure{$Bikey_{seq}$}\Comment{Bikeys of the typing word}
        \Require{$State_{SS}$}\Comment{States during typing}
        \State $NewState_{SS}\gets{\{\}}$
        \If{\textcolor{blue}{N-gram LM}}
        \State \textcolor{blue}{$G\gets{G_{ngram}}$}
        \ElsIf{\textcolor{red}{NN-LM}}
        \State \textcolor{red}{$G_{NLM}\gets{ExtendLM(NLM, C_{input}})$}
        \State \textcolor{red}{$G\gets{G_{NLM}}$}
        \EndIf{}
        \For{$S_{C \circ L}, S_G\ in\ State_{SS}$}
            \State $S\gets{Compose(C \circ L, G, S_{C \circ L}, S_G)}$
            \State $NewState_{SS}\gets{NewState_{SS} + S}$
        \EndFor
        \State $Prune(NewState_{SS}, Bikey_{seq})$
        \State $State_{SS}\gets{NewState_{SS}}$
    \end{algorithmic}
\end{algorithm}

NSS presents three key challenges: handling out-of-vocabulary words (OOV) given NN-LM symbol table constraint, preventing search space exploding caused by assuming word separation at each touch frame, and controlling latency considering dynamic NN-LM inference and on-the-fly FST conversion. We address these through carefully generated FST structure design, accurate pruning strategies, and data structure optimizations.


We conducted extensive live experiments on US English, British English, Spanish in Spain and the US, Portuguese in Portugal. Our key metrics are Words Modified Ratio (WMR), approximating word error rate by reporting the proportion of words modified by the user after their initial commit, and typing speed measured by Words Per Minute (WPM). Online experiment results demonstrated WMR improvement in [0.26\%, 1.19\%], at an acceptable level of latency increase [17\%, 28\%].


The contributions of this work can be summarized as follows:
\begin{itemize}[itemsep=0pt,topsep=2pt,parsep=0pt]
    \item We propose Neural Search Space, integrating the long context representation ability of NN-LM into a carefully designed FST.
    \item We resolve practical problems such as OOV, word separation hypothesis, and latency problems through efficient FST structure design, accurate pruning strategies and data structure optimizations.
    \item We demonstrate the effectiveness of NSS under production environment over millions of users through live experiments, improving the user experience by reducing WMR and enhancing typing speed in a system optimized over decades.
\end{itemize}


\section{Background}





Recent advances of Neural Network LMs(NN-LM), notably projects such as GPT-4 \citep{openai2023gpt}, PaLM 2 \citep{anil2023palm}, demonstrate their superior performance compared to N-gram LMs, particularly in capturing longer context.

Federated Learning (FL) \cite{mcmahan2017communication, kairouz2021advances} with Differential Privacy (DP) \cite{dwork2006calibrating, dwork2014algorithmic} enables Gboard to improve LM quality with user data while preserving user privacy by distributing model training across user devices instead of collecting data centrally. Prior work employed FL to train LMs for Next Word Prediction, Smart Compose, and On-The-Fly rescoring in Gboard following \citet{hard2018federated, xu2023federated}. However, these applications either operate on first pass decoding results produced by N-gram LMs, or do not affect decoding suggestion which has the largest impact on typing experience.

To benefit from FL of NN-LM and retain decoding efficiency, previous research has explored projecting or approximating NN-LMs onto N-gram LMs \citep{chen2019federated, suresh2019distilling, suresh2021approximating}, and making the FST differentiable \citep{hannun2020differentiable}. However, such conversions inevitably incur losses due to limited context and back-off smoothing necessitated by sparsity \cite{chen1999empirical}.




In this work, we replace the N-gram LM within the search space with an NN-LM trained via FL, enhancing long context capabilities. The deployed NN-LM is an LSTM / CIFG model similar to those in \citet{hard2018federated, xu2023federated}.

\section{Challenges}
\label{sec:challenges}


Ideally, an NN-LM would score all known words for optimal coverage. However, vocabulary size is limited due to the high computational cost of the final dense layer. Our deployed NN-LM has a 30k-word vocabulary (top words from Federated Counting), while the full lexicon contains 170k words. Scoring the remaining 140k words in our generated FST poses a key challenge.



Missing the space key and mistyping it with the “cvbn” keys are the two common and consequential mistakes in mobile typing, turning multiple words into one single string (See \cref{app:multiword_demo} for demo cases). Converting <word, probability> pairs to an FST for the current context would only provide NN-LM scores for the first word in such cases, with subsequent words receiving context-less unigram scores. This penalizes and perhaps suppresses multi-word candidates. We address this using dynamic inference in \cref{sec:dynamic_inference}.


Gboard operates under strict latency constraints. Key presses should trigger visible feedback within 20ms as highlighted in \citet{ouyang2017mobile}. NSS inevitably increases latency due to NN-LM inference and FST conversion. Dynamic inference, employed to address space substitution issues, significantly expands the search space by hypothesizing word separations at each frame, further exacerbates this challenge.



\section{Methods}


We detail the $UpdateLM$ and $ExtendLM$ described in \cref{alg:init_search_space} and \cref{alg:extend_search_space} respectively below.

\subsection{Algorithms}

The pseudocode for $UpdateLM$ and $ExtendLM$ is provided in \cref{alg:update_lm} and \cref{alg:extend_lm}.

In $UpdateLM$, $G_{NLM}$ is first set to the initial structure $G_{base}$ (\cref{fig:init_fst}), either by direct reset in decoder initialization or via $ResetFST$. As $G_{base}$ is as large as the full 170k-word vocabulary, and the modified FST will have thousands of new states and arcs on top of that, in-place reset is more efficient than copy. We propose a more compact data structure for efficient reset in  \cref{sec:frequently_modified_fst}.

\begin{algorithm}[H]
    \caption{UpdateLM}
    \label{alg:update_lm}
    \begin{algorithmic}
        \Ensure{$C_{input}$}\Comment{Committed words}
        \Ensure{$NLM$}\Comment{Neural Network LM}
        \Require{$G_{NLM}$}\Comment{Runtime generated FST}
        \If{$G_{NLM} = null$}
            \State $G_{NLM}\gets{G_{base}}$
        \Else
            \State $ResetFST(G_{NLM})$
        \EndIf
        \State $S_{start}\gets0$
        \State $ModifyFST(G_{NLM}, C_{input}, S_{start}, NLM)$
    \end{algorithmic}
\end{algorithm}

Next, $ModifyFST$ inserts the $NLM$ outputs into $G_{NLM}$ as arcs attached on the start state 0 ( \cref{fig:update_lm_fst}).

\begin{algorithm}[H]
    \caption{ExtendLM}
    \label{alg:extend_lm}
    \begin{algorithmic}
        \Ensure{$C_{input}$}\Comment{Committed words}
        \Ensure{$NLM$}\Comment{Neural Network LM}
        \Require{$G_{NLM}$}\Comment{Runtime generated FST}
        \State $S_{extend}\gets{FindStatesToExpand()}$
        \State $DynamicInferencePruning(S_{extend})$
        \For{$S\ in\ S_{extend}$}
            \State $W\gets{FindAdditionalContext(S)}$
            \State $C_{extend}\gets{C_{inputs} + W}$
            \State $ModifyFST(G_{NLM}, C_{extend}, S, NLM)$
        \EndFor
    \end{algorithmic}
\end{algorithm}

Similarly, $ExtendLM$ modifies $G_{NLM}$ at other states chosen dynamically based on context and scores (discussed in \cref{sec:dynamic_inference_pruning}). An example FST structure after $ExtendLM$ is shown in \cref{fig:extend_lm_fst}.

\subsection{FST Structure}

\begin{figure}[H]
    \centering
    \vspace{-0.2cm}
    \includegraphics[width=7cm]{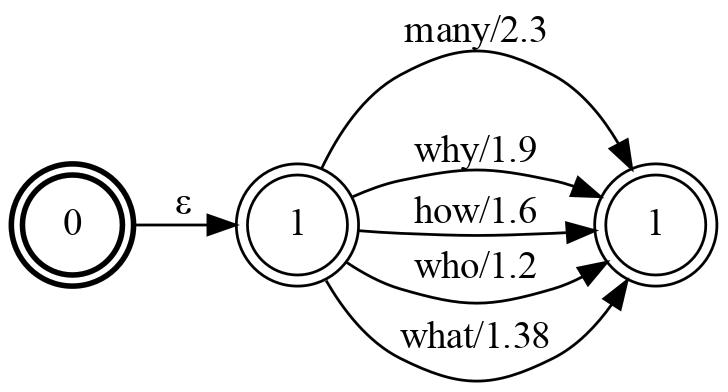}
    \caption{Initial FST Structure}
    \label{fig:init_fst}
    \vspace{-0.2cm}
\end{figure}

The initial structure of the FST in NSS is shown in \cref{fig:init_fst}. State 0 is the start state and state 1 is the unigram state. Unigrams are attached to the unigram state as arcs with format “word/weight”, where weight is the negative log probability. This example only has 5 unigrams. For clarity, we decouple the self-loop on the unigram state by duplicating the unigram state in the graph. Only one zero weight epsilon arc is attached to the start state before any modification.

\begin{figure}[H]
    \centering
    \vspace{-0.2cm}
    \includegraphics[width=7cm]{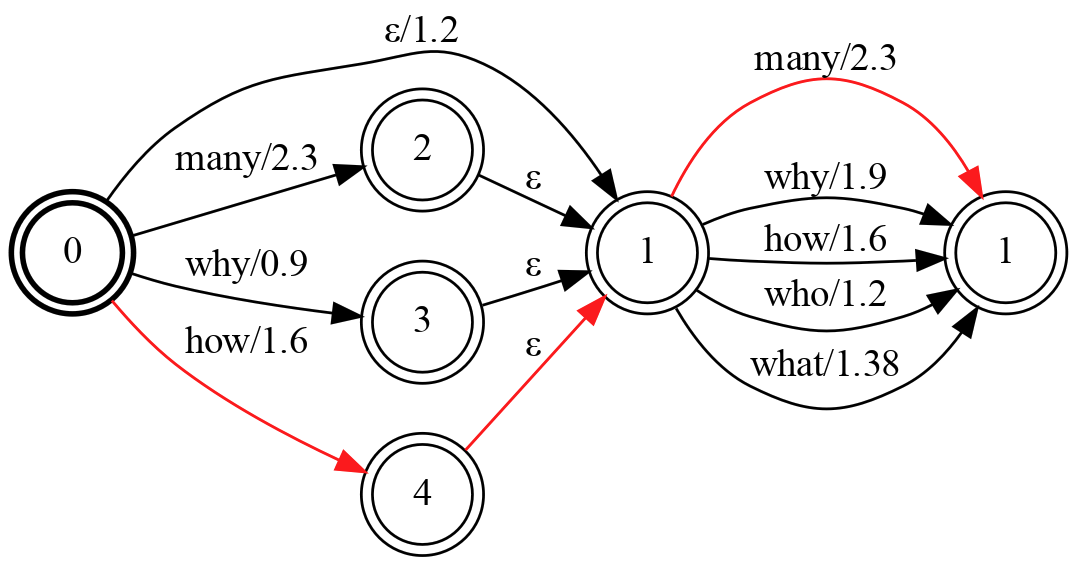}
    \caption{FST Structure after $UpdateLM$, three words are in the NN-LM vocab}
    \label{fig:update_lm_fst}
    \vspace{-0.2cm}
\end{figure}

Given new context, $UpdateLM$ inserts NLM outputs into $G_{NLM}$ as arcs (\cref{fig:update_lm_fst}). Three words and weights are attached to the start state as arcs, each leading to a new state with an epsilon arc to the unigram state. The NN-LM contains fewer words than the total unigrams. The epsilon arc from the start state has the <UNK> probability from the NN-LM. 

The epsilon arc plays a key role in handling OOV: if a word is not in the vocabulary of the NN-LM, the search traverses the epsilon arc to the larger unigram state. Here "OOV" means words in the unigrams but not in the NN-LM; real OOV words are handled by literal decoding and dynamic models following \citet{ouyang2017mobile}, which is the same for N-gram LMs as for NN-LM.

This structure also handles the words with space substitution errors: the first word of the contiguous multi-word candidate is scored by the NN-LM, and the rest receive unigram scores. For example, in \cref{fig:update_lm_fst}, the path of “how many” from state 0 to state 4 to state 1 is highlighted in red.

To provide NN-LM scores for all words in contiguous multi-word candidates, we introduce Dynamic Inference below.

\subsection{Dynamic Inference}
\label{sec:dynamic_inference}

To be able to provide NN-LM scores for all words in contiguous multi-word candidates, the FST structure is expanded dynamically based on the most likely target words users are typing. Inference will run on the concatenation of the base context and the possible target words, and the result probability distribution will be merged into the runtime generated $G_{NLM}$. This process is named Dynamic Inference, which is exactly the $ExtendLM$ in \cref{alg:extend_lm}.

\begin{figure}
    \vspace{-0.3cm}
    \centering
    \includegraphics[width=7.5cm]{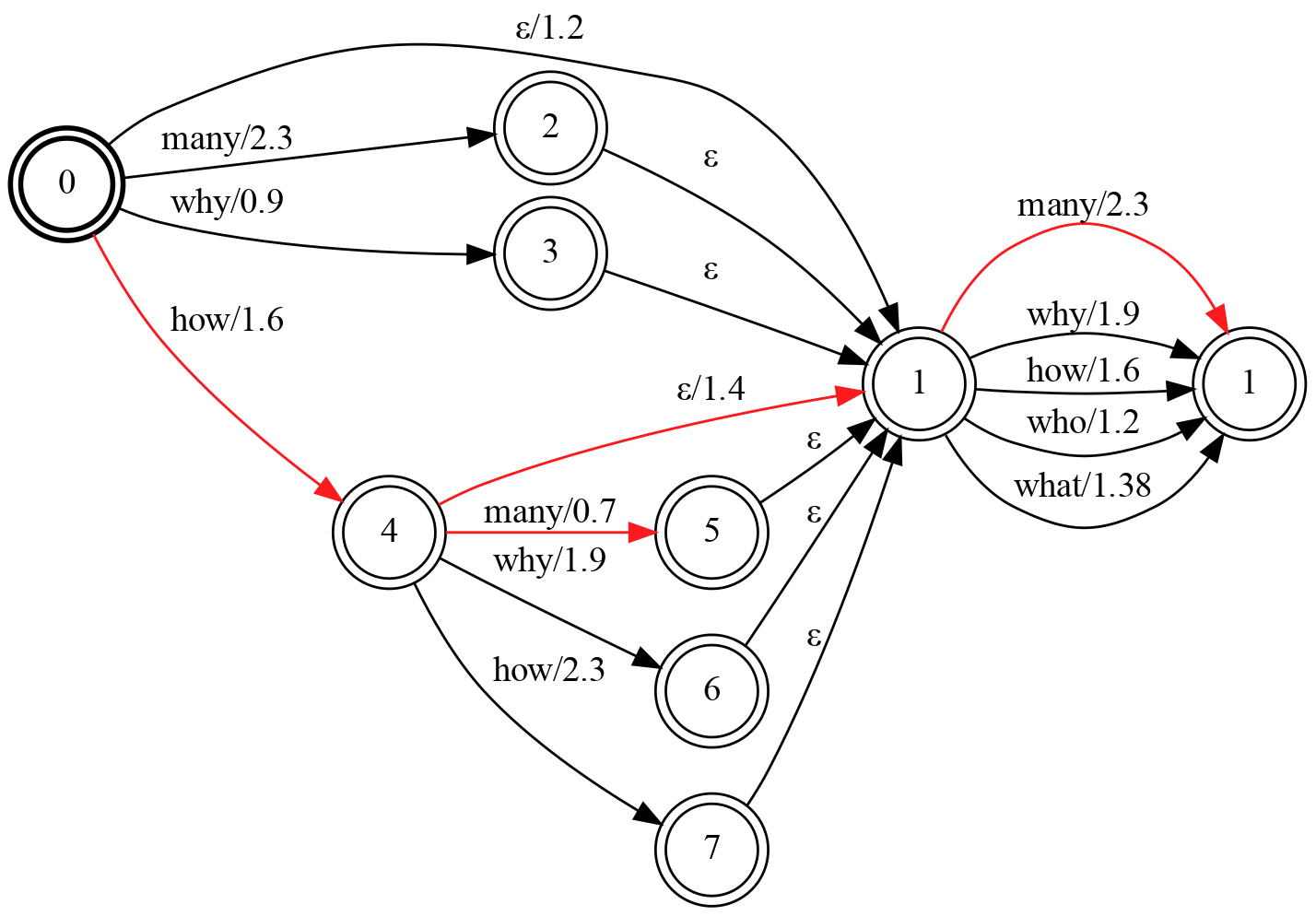}
    \caption{FST structure after one-time dynamic inference}
    \label{fig:extend_lm_fst}
    \vspace{-0.3cm}
\end{figure}

\cref{fig:extend_lm_fst} illustrates an example of dynamic inference. Assuming $FindStatesToExpand()$ returns state $4$, then “how” is the target word, NN-LM inference is conducted on context + “how”, and the outputs are converted to the arcs attached to state $4$, which is very similar to the operations on state 0. Dynamic inference will keep updating the FST at the newly-added states in a recursive manner.

After expansion, there are two paths for “how many”, $0\to4\to5$ and $0\to4\to1\to1$, the former path can provide pure NN-LM scores for the candidate while the latter still provides the mixed scores. The search phase will return the path with higher score, for this case, $0\to4\to5$ path will be returned.

Each state expansion necessitates both NN-LM inference and FST structure updates, leading to a substantial latency increase. \cref{sec:dynamic_inference_pruning} mitigates the number of expansions while \cref{sec:frequently_modified_fst} adapts the FST data structure to frequent modifications efficiently.

\subsection{Latency Optimization}

Various optimizations are explored to meet Gboard's latency requirements. The most effective methods are listed below.

\subsubsection{Arc Pruning}

Each $G_{NLM}$ modification involves inserting 30k <word, score> pairs. However, since lower scores are unlikely to survive in beam search phase, words with probabilities less than a fixed $T_{arc}$ are omitted from the FST. $T_{arc}$ is set to $e^{-15}$ for $UpdateLM$ and $e^{-12}$ for $ExtendLM$. This generally keeps only 1k to 5k words, which reduced the tap typing latency increment from +211.54\% to +81.51\% in offline evaluation.

Settings and detailed results of the offline evaluation can be seen in \cref{app:offline_setting} and \cref{app:arc_pruning_offline_results} respectively.

\subsubsection{Dynamic Inference Pruning}
\label{sec:dynamic_inference_pruning}

Ideally, we should expand $G_{NLM}$ at all states if possible, however, the time complexity is an unbearable $O(N^L)$, where N is the number of words in NN-LM and L is the length of the candidate.

Dynamic Inference Pruning is applied in \cref{alg:extend_lm} to reduce the complexity. We adopt two rules simultaneously to decide whether a state should be expanded.

\begin{itemize}[itemsep=0pt,topsep=2pt,parsep=0pt]
    \item Only states with scores larger than a threshold $T_{extend}$ are eligible for expansion.
    \item Only the top $N$ states with eligible scores may be expanded.
\end{itemize}

We explored various thresholds in offline evaluation, empirically choosing $T_{extend} = e^{-12}$ and $N = 1$, which increases tap typing latency by 79.92\%.



\subsubsection{Frequently Modified FST}
\label{sec:frequently_modified_fst}

The default mutable FST implementation we use is OpenFST \citep{allauzen2007openfst}, in which arcs are stored independently per state to offer flexibility to add and remove arcs and states. However, it's inefficient when the FST is incrementally updated and frequently reset. Using reset as an example, we would need to delete the arcs of each state first and then delete the states, which is expensive.

Based on this requirement for incremental updates and frequent resets, we propose a customized FST implementation. The arcs of all states are stored in the same array, and the FST maintains a map from states to the indices of their corresponding arcs in the large array. When resetting the FST, we only need to clear the single array of arcs and then delete the map of states; the arc array can be reused instead of reallocated each time.

The FST structures are illusrated in figures in \cref{app:customized_fst_struct}.

\begin{table*}[ht]
    \centering
    \begin{tabular}{cccccc}
        \hline
        Language & Devices & WMR(\%) & WPM(\%) & Latency(\%) & Total Users(M) \\
        \hline
        en-US & ALL & \textbf{-0.26} & +0.06 & +24.13 & 14.00 \\
        \hline
        es-ES & ALL & \textbf{-0.77} & \textbf{+0.40} & +23.13 & 3.30 \\
        \hline
        pt-PT & ALL & \textbf{-0.99} & +0.59 & +28.34 & 0.85 \\
        \hline
        \multirow{3}{*}{en-GB} & ALL & \textbf{-1.19} & +0.40 & +17.92 & 1.99 \\
        & 3+ GB & \textbf{-1.31} & +0.48 & +13.98 & 1.43 \\
        & 6+ GB & \textbf{-1.13} & +0.30 & +7.36 & 0.64 \\
        \hline
        \multirow{3}{*}{es-US} & ALL & \textbf{-1.03} & \textbf{+0.39} & +17.43 & 14.37 \\
        & 3+ GB & \textbf{-1.24} & \textbf{+0.43} & +15.14 & 8.86 \\
        & 6+ GB & \textbf{-1.24} & \textbf{+0.52} & +12.89 & 1.46 \\
        \hline
    \end{tabular}
    \caption{Live Experiment results for en-US, es-ES, pt-PT, en-GB and es-US.}
    \label{tab:le_results}
\end{table*}

\section{Evaluation}

We conducted live experiments on uniform random samples of the eligible Gboard populations \cite{sivek2022spatial} for US English (en-US), GB English (en-GB), ES Spanish (es-ES), PT Portuguese (pt-PT) and US Spanish (es-US).

\subsection{A/B Metrics}

Metrics in the A/B experiments to measure the quality and latency are:

\begin{itemize}[itemsep=0pt,topsep=2pt,parsep=0pt]
    \item \textbf{Words Modified Ratio (WMR)}: The ratio of words being modified during typing or after committed; improvement is shown by reduction.
    \item \textbf{Words Per Minute (WPM)}: The number of committed words per minute.
    \item \textbf{Latency}: The average time for decoding.
    \item \textbf{Total Users}: The number of users participating in the experiments with the target languages.
\end{itemize}

\subsection{Experiment Setup}

There are two arms in the live experiments:
\begin{itemize}[itemsep=0pt,topsep=2pt,parsep=0pt]
    \item \textbf{Control Arm}: the LM is a N-gram FST which is obtained by approximating the NN-LM trained via Federated Learning or counting from server corpus.
    \item \textbf{NSS Arm}: the LM is a simple FST generated by one-layer LSTM at runtime.
\end{itemize}


The NN-LM in live experiment is a one-layer LSTM model with the following configuration:

\begin{itemize}[itemsep=0pt,topsep=2pt,parsep=0pt]
    \item $vocab\_size = 30k$
    \item $embedding\_dim = 96$
    \item $lstm\_size = 670$
    \item $total\_parameters = 6.4M$
    \item $training\_loss$: cross entropy of next word prediction.
\end{itemize}

We hypothesize that the quality of NSS is limited by the latency. To verify this, we also report metrics restricted to high-end devices with memory larger than 3G / 6G for en-GB and es-US.

\subsection{Result Analysis}

The online live experiment results are listed in \cref{tab:le_results}. All metrics other than Total Users are reported as percent changes relative to the control arm. All latency changes and all bolded WMR and WPM changes are significant at a 0.05 level (null hypothesis: the metric change is 0).

It's observed in \cref{tab:le_results} that:

\begin{itemize}[itemsep=0pt,topsep=2pt,parsep=0pt]
    \item The NSS arm reduces WMR with a 95\% confidence interval of [0.26\%, 1.19\%] on various languages while increasing latency in the [17\%, 28\%] range.
    \item Higher-end devices exhibits marginally greater improvements in WMR and WPM with smaller latency increases. Specifically for example, for es-US, the latency increment on 6G+ devices is 12.89\%, 5\% less than the increment in ALL devices, while the WMR and WPM improvement are larger. The metrics of en-GB on 6G+ devices are not better than the other settings, which we argue is potentially caused by the small population attending the live experiments.
    \item WPM is consistently improved, suggesting that the latency cost does not adversely affect user experience.
\end{itemize}


From the perspective of production, as a well optimized system, WMR of Gboard decoder ranges within [4.5\%, 7\%] for locales studied in this paper, the relative improvement larger than 0.1\% is deemed significant. Additionally, the observed latency increase fell below the sensitive threshold, allowing us to achieve both WMR/WPM optimization and latency control. Specifically, the Gboard decoder consists of a series of modules, with NSS positioned upstream. Any latency increase in NSS can potentially impact downstream modules like key correction and auto-corrections, negatively affecting key metrics. Our experimental findings indicate that if the latency increase remains below a certain level, downstream modules are unaffected, and users do not perceive the latency change.

The confirmation on the hypothesis regarding higher-end devices empowers us to deploy more powerful models on devices with greater capabilities, further enhancing the user experience.

It's expected that the quality improvement on en-US is much smaller than the other locales. Due to the high product priority, the baseline of en-US is much stronger than other locales, which adopts the FST approximated from a FL trained LM \citep{chen2019federated} and a specialized N-gram model for the search domain.


\section{Discussions}



This work successfully bridged the gap between the long range context awareness power of NN-LMs and the efficiency requirements of Gboard's decoder. By creatively adapting NN-LMs to an FST structure and implementing latency optimizations, we deployed the Neural Search Space in production. Experiments demonstrated that NSS significantly improves decoding quality, particularly on higher-end devices, with an acceptable latency trade-off. This direct integration unlocks potential for future enhancements driven by NN-LM advancements, promising further gains in keyboard decoding and overall user experience.

Building upon NSS, we identify several promising directions for future research:

\begin{itemize}[itemsep=0pt,topsep=2pt,parsep=0pt]
    \item \textbf{Integrating Transformer models}: Transformer models are known for their superior quality and training efficiency.  Exploring their integration within NSS presents an exciting opportunity. However, a key challenge here is maintaining system performance, given the substantial computational resources and memory required by the Transformer model to handle long contexts. Further investigation is needed to assess the quality gains achievable with models constrained to fewer than 10 million parameters due to these system limitations.
    \item \textbf{Leveraging richer context}: Compared to traditional n-gram models, NN-LMs offer a more flexible framework for incorporating diverse contextual information.  Integrating signals like application domain, country, time, and extended user history holds the potential to further enhance model quality.
    \item \textbf{Exploring SentencePiece LMs}: The current NSS utilizes word-level LMs, which can be limited by OOV issues. Employing SentencePiece \citep{kudo2018subword} LMs could improve performance by providing better word coverage and a more nuanced representation of language.
\end{itemize}

Beyond enhancing NSS, another avenue for exploration is replacing the FST-based decoder with a neural decoder. While we have investigated this approach, certain challenges hinder its immediate adoption as a full replacement:

\begin{itemize}[itemsep=0pt,topsep=2pt,parsep=0pt]
    \item \textbf{Quality Gap}: The current system, refined over a decade by numerous engineers, incorporates extensive prior knowledge about error patterns, which is difficult to encapsulate within a single neural model.
    \item \textbf{Debugging Challenges}: The current system allows for straightforward debugging and correction of errors by adjusting weights in FSTs. Transitioning to a purely neural decoder would sacrifice this flexibility..
\end{itemize}

However, we recognize the inherent advantages of neural models, such as superior semantic understanding and context capture. Therefore, we continue to experiment with end-to-end approaches. One promising avenue is to run neural models in parallel with the existing system and merge their candidate suggestions, leveraging the strengths of both FSTs and neural approaches. This hybrid approach allows us to benefit from the  precision and debuggability of the FST-based system while capitalizing on the advanced contextual understanding of neural models.


\bibliography{main}


\appendix
\section{Appendix}
\label{sec:appendix}

\subsection{Multi-word Demo}
\label{app:multiword_demo}

\begin{figure*}[ht]
    \centering
    \includegraphics[width=12cm]{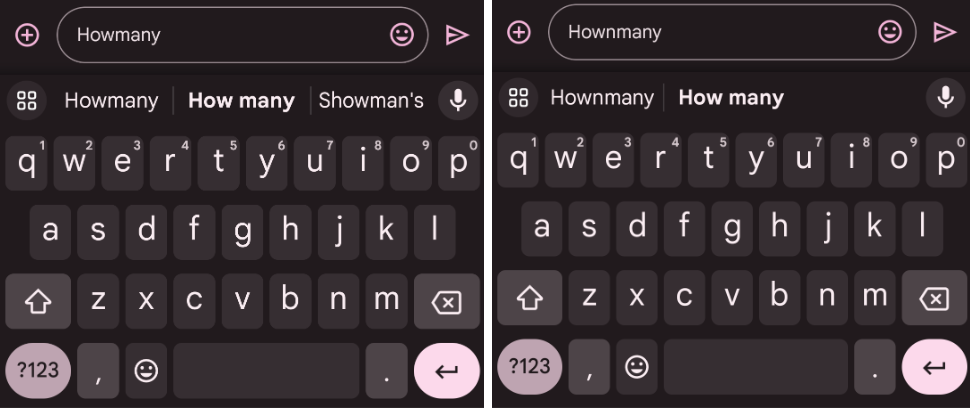}
    \caption{Word with Space Substitution Errors. Left: misses the space key. Right: mistypes the space with "n" ("cvbn")}
    \label{fig:multi_word_problem}
\end{figure*}

The two frequently multi-word typos described in \cref{sec:challenges} can be seen in \cref{fig:multi_word_problem}.

\subsection{Offline Evaluation Setting}
\label{app:offline_setting}

TypingTester is the tool used for offline evaluation. It's a testing framework for the Gboard decoder and related C++ code. It repeatedly runs the decoder over a sequence of touch points, compares the output text to expected text, and estimates metrics like word error rate, next word prediction accuracy, decode time, and more.

The dataset used for offline evaluation contains touch points for 2500 sentences, which is gathered from 50 volunteers by typing the same 50 sentences.

In this paper, typingtester is used to report the relative decoding latency change, the tests run on a workstation with 3.7Ghz, 6 core Intel CPU.

\subsection{Arc Pruning Latency Impact}
\label{app:arc_pruning_offline_results}

The latency impact of arc pruning with different thresholds is shown in \cref{tab:latency_arc_pruning}. We report the relative change of latency over the prod N-gram search space, the latencies for tap-typing and gesture typing are reported separately.

\begin{table}[ht]
    \centering
    \begin{tabular}{cc|cc}
        \hline
        $T_{Update}$ & $T_{Extend}$ & Tap (\%) & Gesture (\%) \\
        \hline
        $e^{-15}$ & $e^{-12}$ & +81.51 & -9.28\\
        \hline
        $e^{-10}$ & $e^{-10}$ & +57.96 & -36.03 \\
        $e^{-12}$ & $e^{-12}$ & +79.85 & -25.37 \\
        $e^{-15}$ & $e^{-15}$ & +138.36 & -3.21 \\
        $0$ & $0$ & +211.54 & +15.59 \\
        \hline
    \end{tabular}
    \caption{Latency change on various arc threshold combinations}
    \label{tab:latency_arc_pruning}
\end{table}

The chosen thresholds for Neural Search Space are $e^{-15}$ for $UpdateLM$ and $e^{-12}$ for $ExtendLM$. The latency increase of tap typing is 81.51\%, which is larger than the latency increment online due to the reasons listed below.

\begin{itemize}[itemsep=0pt,topsep=2pt,parsep=0pt]
    \item The optimization gap between workstation and phone devices, eg: tflite inference is faster on device.
    \item Not all modules are involved in the offline evaluation.
\end{itemize}

The gesture latency is reduced by 9.28\%, which benefits from the inherent property that gesture typing is free of the missing/mistyping space key problem, such that the dynamic inference defined in \cref{alg:extend_lm} is not required.

No pruning happens when the thresholds are set to 0, and 130.03\% and 24.87\% latency savings on tap typing and gesture typing are observed respectively comparing to the chosen thresholds.

\subsection{Dynamic Inference Pruning Latency Impact}
\label{app:dynamic_inference_pruning_latency_offline_results}

Dynamic Inference pruning strategy contains two thresholds, $N$ controls how many states could be expanded at most per char and $T_{expand}$ controls whether a state is eligible to be expanded, which are set to be $1$ and $e^{-12}$ respectively after verifying on live experiments.

\cref{tab:latency_dynamic_inference_pruning} displays the latency impact of various thresholds. Gesture latency is not affected significantly as it doesn't have dynamic inference. Tap latency is affected by the range from 10\% to 20\% when changing the thresholds. The latency increase is reduced from 79.92\% to 40.86\% if cancelling the dynamic inference, which defines the loose upper bound of latency increase in dynamic inference optimization.

\begin{table}[ht]
    \centering
    \begin{tabular}{cc|cc}
        \hline
        $N$ & $T_{extend}$ & Tap (\%) & Gesture (\%) \\
        \hline
        1 & $e^{-12}$ & +79.92 & -7.60 \\
        \hline
        2 & $e^{-12}$ & +94.51 & -8.23 \\
        3 & $e^{-12}$ & +99.31 & -8.35 \\
        1 & $e^{-15}$ & +91.17 & -6.19 \\
        1 & $e^{-10}$ & +69.21 & -7.07 \\
        \multicolumn{2}{c|}{No Dynamic Inference} & +40.86 & -8.45 \\
        \hline
    \end{tabular}
    \caption{Latency change on various dynamic inference thresholds}
    \label{tab:latency_dynamic_inference_pruning}
\end{table}

\subsection{Customized FST structure}
\label{app:customized_fst_struct}

\begin{figure}[ht]
    \centering
    \includegraphics[width=5.8cm]{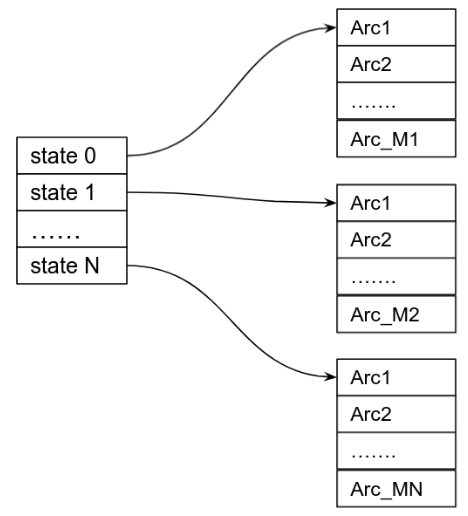}
    \caption{Default Modifiable FST}
    \label{fig:fst_default}
\end{figure}

\begin{figure}[ht]
    \vspace{-0.2cm}
    \centering
    \includegraphics[width=5.8cm]{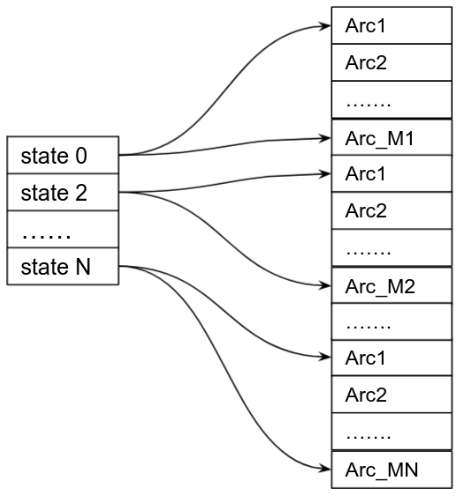}
    \caption{Optimized Modifiable FST in incremental update scenarios}
    \label{fig:fst_optimized}
\end{figure}

As described in \cref{sec:frequently_modified_fst}, the default modified FST implementation in OpenFST is illustrated in \cref{fig:fst_default}. while the customized implementation of a frequently updated FST is illustrated in \cref{fig:fst_optimized}.

\end{document}